\documentclass[review]{elsarticle}

\usepackage{lineno,hyperref}
\nolinenumbers
\usepackage{microtype}
\usepackage{graphicx}
\usepackage{subfigure}
\usepackage{amsmath}
\usepackage{amssymb}
\newtheorem{thm}{Theorem}
\usepackage{graphicx}
\usepackage{booktabs} 

\usepackage{hyperref}

\nolinenumbers
\journal{Pattern Recognition}

\begin{document}

\begin{frontmatter}

\title{Online Anomaly Detection in Surveillance Videos with Asymptotic Bounds on False Alarm Rate}

\author{Keval Doshi, Yasin Yilmaz}\corref{mycorrespondingauthor}
\address{University of South Florida\\4202 E Fowler Ave, Tampa, FL 33620}

% %% or include affiliations in footnotes:
% \author[mymainaddress,mysecondaryaddress]{Elsevier Inc}

% \author[mysecondaryaddress]{Global Customer Service\corref{mycorrespondingauthor}}
\cortext[mycorrespondingauthor]{Corresponding author}

% \address[mymainaddress]{1600 John F Kennedy Boulevard, Philadelphia}
% \address[mysecondaryaddress]{360 Park Avenue South, New York}

\begin{abstract}
Anomaly detection in surveillance videos is attracting an increasing amount of attention. Despite the competitive performance of recent methods, they lack theoretical performance analysis, particularly due to the complex deep neural network architectures used in decision making. Additionally, online decision making is an important but mostly neglected factor in this domain. Much of the existing methods that claim to be online, depend on batch or offline processing in practice. Motivated by these research gaps, we propose an online anomaly detection method in surveillance videos with asymptotic bounds on the false alarm rate, which in turn provides a clear procedure for selecting a proper decision threshold that satisfies the desired false alarm rate. Our proposed algorithm consists of a multi-objective deep learning module along with a statistical anomaly detection module, and its effectiveness is demonstrated on several publicly available data sets where we outperform the state-of-the-art algorithms. All codes are available at \url{https://github.com/kevaldoshi17/Prediction-based-Video-Anomaly-Detection-}.
\end{abstract}
\begin{keyword}
\texttt computer vision; video surveillance; anomaly detection; asymptotic performance analysis; deep learning; online detection

\end{keyword}

\end{frontmatter}

\section{Introduction}
\label{intro}
The rapid advancements in the technology of closed-circuit television (CCTV) cameras and their underlying infrastructural components such as network, storage, and processing hardware have led to a sheer number of surveillance cameras implemented all over the world, and estimated to go beyond 1 billion globally, by the end of the year 2021 \cite{videosurveillance}. Video surveillance is an essential tool used in law enforcement, transportation, environmental monitoring, etc. mainly for improving security and public safety. For example, it has become an inseparable part of crime deterrence and investigation, traffic violation detection, and traffic management. However, considering the massive amounts of videos generated in real-time, manual video analysis by human operator becomes inefficient, expensive, and nearly impossible, which in turn makes a great demand for automated and intelligent methods for analyzing and retrieving important information from videos, in order to maximize the benefits of CCTV.

\iffalse

\fi

One of the most important, challenging and time-critical tasks in automated video surveillance is the detection of abnormal events such as traffic accidents and violations, crimes, and natural disasters. Hence, video anomaly detection has become an important research problem in the recent years. Anomaly detection in general is a vast, crucial, and challenging research topic, which deals with the identification of data instances deviating from nominal patterns. It has a wide range of applications, e.g., in medical health care\cite{medicAD}, cyber-security \cite{newinfobased}, hardware security \cite{hardwareSecurity}, aviation \cite{nasa1}, and spacecraft monitoring \cite{spacecraft}. \iffalse One of the key challenges of anomaly detection, specifically in video arises from the fact that defining a notion of abnormality that encompasses all possible anomalous data patterns is nearly impossible. \fi

Given the important role that video anomaly detection can play in ensuring safety, security and sometimes prevention of potential catastrophes, one of the main outcomes of a video anomaly detection system is the real-time decision making capability. Events such as traffic accidents, robbery, and fire in remote places require immediate counteractions to be taken in a timely manner, which can be facilitated by the real-time detection of anomalous events. Despite its importance, a very limited body of research has focused on online and real-time detection methods. Moreover, some of the methods that claim to be online heavily depend on batch processing of long video segments. For example, \cite{liu2018future} performs a normalization step which requires the entire video.

A vast majority of the recent state-of-the-art video anomaly detection methods depend on complex neural network architectures \cite{sultani2018real}. Although deep neural networks provide superior performance on various machine learning and computer vision tasks, such as object detection \cite{dai2016r}, image classification \cite{krizhevsky2012imagenet}, playing games \cite{silver2017mastering}, image synthesis\cite{reed2016generative}, etc., where sufficiently large and inclusive data sets are available to train on, there is also a significant debate on their shortcomings in terms of interpretability, analyzability, and reliability of their decisions \cite{jiang2018trust}. For example, \cite{papernot2018deep,sitawarin2019defending} propose using a nearest neighbor-based approach together with deep neural network structures to achieve robustness, interpretability for the decisions made by the model, and as defense against adversarial attack. Additionally, to the best of the our knowledge, none of the neural network-based video anomaly detection methods has been analyzed in terms of performance guarantees. On the other hand, statistical and nearest neighbor-based methods remain popular due to their appealing characteristics such as being amenable to performance analysis, computational efficiency, and robustness \cite{chen2018explaining,gu2019statistical}. 

Motivated by the aforementioned domain challenges and research gaps, we propose a hybrid use of neural networks and statistical $k$ nearest neighbor ($k$NN) decision approach for finding anomalies in video in an online fashion. In summary, our contributions in this paper are as follows:
\begin{itemize}
    \item We propose a novel framework composed of deep learning-based feature extraction from video frames, and a statistical sequential anomaly detection algorithm.
    \item We derive an asymptotic bound on the false alarm rate of our detection algorithm, and propose a technique for selecting a proper threshold which satisfies the desired false alarm rate.
    \item We extensively evaluate our proposed framework on publicly available video anomaly detection data sets.
\end{itemize}

The remainder of the paper is organized as: Related Work (Section \ref{related}), Proposed Method (Section \ref{proposed}), Experiments (Section \ref{experiments}), and Conclusion (Section \ref{conclusion}). 

\section{Related Work}
\label{related}
Semi-supervised detection of anomalies in videos, also known as outlier detection, is a commonly adopted learning technique due to the inherent limitations in availability of annotated and anomalous instances. This category of learning methods deals with learning a notion of normality from nominal training videos, and attempts to detect deviations from the learned normality notion. \cite{cheng2015video,ionescu2019object}. There are also several supervised detection methods, which train on both nominal and anomalous videos. The main drawback of such methods is the difficulty in finding  frame-level labeled, representative, and inclusive anomaly instances. To this end, \cite{sultani2018real} proposes using a deep multiple instance learning (MIL) approach to train on video-level annotated videos, in a weakly supervised manner. Although training on anomalous videos would enhance the detection capability on similar anomaly events, supervised methods typically suffer from unknown and novel anomaly types.

One of the key components of the video anomaly detection algorithms is the extraction of meaningful features, which can capture the difference between the nominal and anomalous events within the video. The selection of feature types has a significant impact on the identifiability of types of anomalous events in the video sequences. Many early video anomaly detection techniques and some recent ones focused on the trajectory features \cite{anjum2008multifeature}, which limits their applicability to the detection of the anomalies related to the trajectory patterns, and moving objects. For instance, \cite{fu2005similarity} studied detection of abnormal vehicle trajectories such as illegal U-turn. \cite{morais2019learning} extracts human skeleton trajectory patterns, and hence is limited to only the detection of abnormalities in human behavior.

Motion and appearance features are another class of widely used features in this domain. \cite{saligrama2012video} extracts motion direction and magnitudes, to detect spatio-temporal anomalies. Histogram of optical flow \cite{chaudhry2009histograms,colque2016histograms}, and histogram of oriented gradients \cite{dalal2005histograms} are some other commonly used hand-crafted feature extraction techniques used in the literature. Sparse coding based methods \cite{zhao2011online} are also applied in detection of video anomalies. They learn a dictionary of normal sparse events, and attempt to detect anomalies based on the reconstructability of video from the dictionary atoms. \cite{mo2013adaptive} uses sparse reconstruction to learn joint trajectory representations of multiple objects. %Another class of Due to successful demonstration

In contrary to the hand-crafted feature extraction, are the neural network based feature learning methods. \cite{xu2015learning} learns the appearance and motion features by deep neural networks. \cite{luo2017remembering} utilizes Convolutional Neural Networks (CNN), and Convolutional Long Short Term Memory (CLSTM) to learn appearance and motion features, respectively.
Neural network based approaches have been recently dominating the literature. For example, \cite{ravanbakhsh2017abnormal} trains Generative Adversarial Network (GAN) on normal video frames, to generate internal scene representations (appearance and motion), based on a given frame and its optical flow, and detects deviation of the GAN output from the normal data, by AlexNet \cite{krizhevsky2012imagenet}. \cite{sabokrou2018adversarially} trains a GAN-like adversarial network, in which a reconstruction component learns to reconstruct the normal test frames, and attempts to train a discriminator by gradually injecting anomalies to it, while concurrently the discriminator (detector) learns to detect the anomalies injected by the reconstructor. In \cite{doshi2020continual,doshi2020any}, a transfer learning based approach is used for continual learning for anomaly detection in surveillance videos from a few samples.

\section{Proposed Method}
\label{proposed}

\begin{figure*}[th]
\centering
\includegraphics[width=1\textwidth]{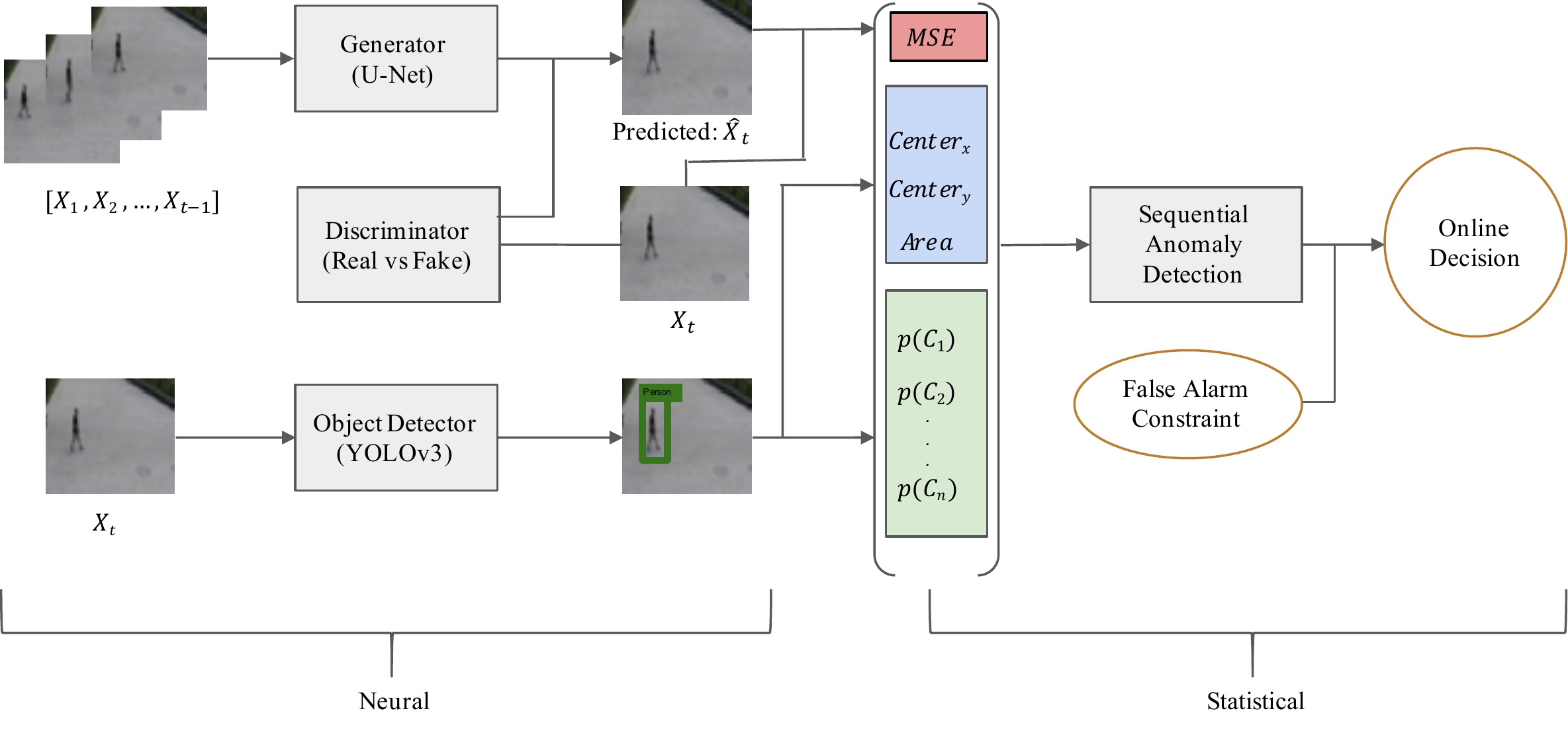}
\vspace{-2mm}
\caption{Proposed MONAD framework. At each time $t$, neural network-based feature extraction module provides motion (MSE), location (center coordinates and area of bounding box), and appearance (class probabilities) features to the statistical anomaly detection module, which automatically sets its decision threshold to satisfy a false alarm constraint and makes online decisions.}
\label{f:system}
\vspace{-2mm}
\end{figure*}

\subsection{Motivation}

Anomaly detection in surveillance videos is defined as the identification of unusual events which do not conform to the expectation. We base our study on two important requirements that a successful video anomaly detector should satisfy: (i) extract meaningful features which can be utilized to distinguish nominal and anomalous data; and (ii) provide a decision making strategy which can be easily tuned to satisfy a given false alarm rate. While existing works partially fulfills the first requirement by defining various constraints on spatial and temporal video features, they typically neglect providing an analytical and amenable decision strategy. Motivated by this shortcoming, we propose a unified framework called Multi-Objective Neural Anomaly Detector (MONAD\footnote{\emph{Monad} is a philosophical term for infinitesimal unit, and also a functional programming term for minimal structure.}). Like \emph{monads} provide a unified functional model for programming, our proposed MONAD unifies deep learning-based feature extraction and analytical anomaly detection by incorporating two modules, as shown in Figure \ref{f:system}. The first module consists of a Generative Adversarial Network (GAN) based future frame predictor and a lightweight object detector (YOLOv3) to extract meaningful features. The second module consists of a nonparametric statistical algorithm which uses the extracted features for online anomaly detection. To the best of our knowledge, this is the first work to present theoretical performance analysis for a deep learning-based video anomaly detection method. Our MONAD framework is described in detail in the following sections.  

\subsection{Feature Selection}

Most existing works focus on a certain aspect of the video such as optical flow, gradient loss or intensity loss. This in turn restrains the existing algorithms to a certain form of anomalous event which is manifested in the considered video aspect. However, in general, the type of anomaly is broad and unknown while training the algorithm. For example, an anomalous event can be justified on the basis of appearance (a person carrying a gun), motion (two people fighting) or location (a person walking on the roadway). To account for all such cases, we create a feature vector $F_t^i$ for each object $i$ in frame $X_t$ at time $t$, where $F_t^i$ is given by $[w_1F_{motion},w_2F_{location},w_3F_{appearance}]$. The weights $w_1, w_2, w_3$ are used to adjust the relative importance of each feature category. 

\subsection{Frame Prediction}
\label{s:predict}

A heuristic approach for detecting anomalies in videos is by predicting the future video frame $\widehat{X}_{t}$ using previous video frames $\{{X}_{1},{X}_{2},\dots,{X}_{t-1}\}$, and then comparing it to $X_t$ through mean squared error (MSE). Instead of deciding directly on MSE, we use MSE of video frame prediction to obtain motion features (Section \ref{s:feature}). GANs are known to be successful in generating realistic images and videos. However, regular GANs might face the vanishing gradient problem during learning as they hypothesize the discriminator as a classifier with the sigmoid cross entropy loss function. To overcome this problem, we use a modified version of GAN called Least Square GAN (LS-GAN) \cite{mao2017least}. The GAN architecture comprises of a generator network $G$ and a discriminator network $D$, where the function of $G$ is to generate frames that would be difficult-to-classify by $D$. Ideally, once $G$ is well trained, $D$ cannot predict better than chance. Similar to \cite{liu2018future}, we employ a U-Net \cite{ronneberger2015u} based network for $G$ and a patch discriminator for $D$. 

For training the generator $G$, we follow \cite{liu2018future}, and combine the constraints on intensity, gradient difference, optical flow, and adversarial training to get the following objective function
\begin{equation}
\begin{split}
     L_G = \gamma_{int} L_{int}(\widehat{X},X) + \gamma_{gd} L_{gd}(\widehat{X},X) +\\ \gamma_{of} L_{of}(\widehat{X},X) + \gamma_{adv} L_{adv}(\widehat{X},X)
\end{split}
\end{equation}
where ${\gamma_{int},\gamma_{gd},\gamma_{of},\gamma_{adv}} \geq 0$ are the corresponding weights for the losses.

\textbf{Intensity loss} is the $l_1$ or $l_2$ distance between the predicted frame $\widehat{X}$ and the actual frame $X$, which is used to maintain similarity between pixels in the RGB space, and given by 
\begin{equation}
L_{int}(\widehat{X},X) = \left \|{\widehat{X}- X}\right \|^2.
\end{equation}

\textbf{Gradient difference loss} is used to sharpen the image prediction and is given by 
\begin{equation}
\begin{split}
L_{gd}(\widehat{X},X) = \sum_{i,j}\left \| | \widehat{X}_{i,j}-\widehat{X}_{i-1,j} | - \left | {X}_{i,j}-{X}_{i-1,j} \right | \right \|_1 \\+ \left \|  | \widehat{X}_{i,j}-\widehat{X}_{i,j-1}  | - \left | {X}_{i,j}-{X}_{i,j-1} \right | \right \|_1
\end{split}
\end{equation}
where $(i,j)$ denotes the spatial index of a video frame.

\textbf{Optical flow loss} is used to improve the coherence of motion in the predicted frame, and is given by
\begin{equation}
L_{of}(\widehat{X}_{t+1},X_{t+1},X_{t}) = \left \|f(\widehat{X}_{t+1},X_{t}) - f(X_{t+1},X_{t})\right \|_1
\end{equation}
where $f$ is a pretrained CNN-based function called Flownet, and is used to estimate the optical flow.

\textbf{Adversarial generator loss} is minimized to confuse $D$ as much as possible such that it cannot discriminate the generated predictions, and is given by 
\begin{equation}
L_{adv}(\widehat{X}) = \sum_{i,j}\frac{1}{2}L_{MSE}(D(\widehat{X}_{i,j}),1)
\end{equation}
where $D(\widehat{X}_{i,j})=1$ denotes ``real" decision by $D$ for patch $(i,j)$, $D(\widehat{X}_{i,j})=0$ denotes ``fake" decision, and $L_{MSE}$ is the mean squared error function.

\subsection{Object Detection}
\label{s:detect}

\begin{figure*}[!htb]
  \includegraphics[width=\linewidth]{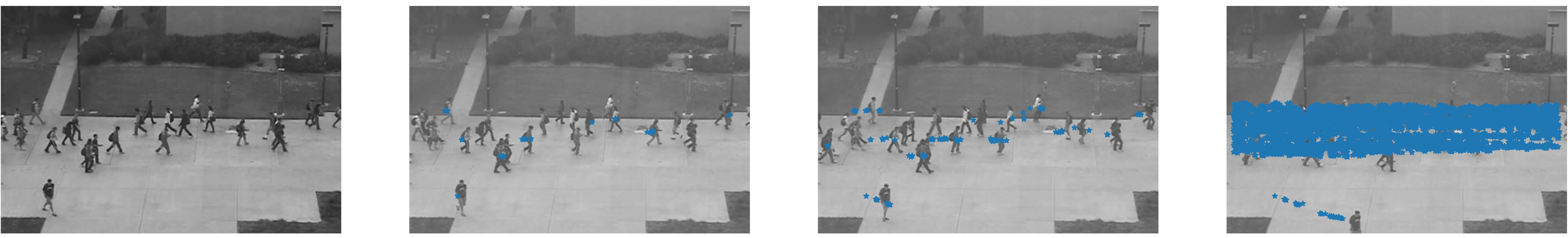}
  \includegraphics[width=\linewidth]{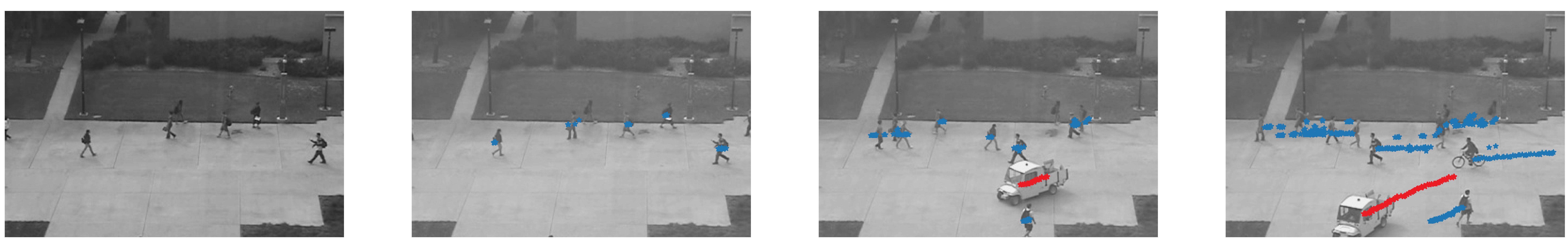}
  \caption{Example video frames from the UCSD Ped2 dataset showing the extraction of bounding box center (location) feature in nominal training data (top row) and test data (bottom row). Columns from left to right correspond to the first, 30th, 150th, and the last frame in all training videos (top row), and in a test video (bottom row). In the test video, the unusual path of golf cart, shown with red dots, together with the class probability and high prediction error (MSE) due to unusual speed of cart statistically contribute to the anomaly decision. Best viewed in color.}
  \label{f:Tracking}
\end{figure*}

We propose to detect objects using a real-time object detection system such as You Only Look Once (YOLO) \cite{redmon2016you} to obtain location and appearance features (Section \ref{s:feature}). 
The advantage of YOLO is that it is capable of processing higher frames per second on a GPU while providing the same or even better accuracy as compared to the other state-of-the-art models such as SSD and ResNet. Speed is a critical factor for online anomaly detection, so we currently prefer YOLOv3 in our implementations. For each detected object in image $X_t$, we get a bounding box (location) along with the class probabilities (appearance). As shown in Fig. \ref{f:Tracking}, we monitor the center of the bounding boxes to track paths different objects might take in the training videos. Instead of simply using the entire bounding box, we monitor the center of the box and its area to obtain location features. This not only reduces the complexity, but also effectively avoids false positives in case the bounding box is not tight. In a testing video, objects diverging from the nominal paths and class probabilities will help us detect anomalies, as explained in Section \ref{s:anomaly}.

\subsection{Feature Vector}
\label{s:feature}

Finally, for each object $i$ detected in a frame, we construct a feature vector as: 
% \begin{center}
\begin{equation}
F_t^i =  
\begin{bmatrix}
\vspace{-5mm}
\\ w_1MSE(X_t,\widehat{X}_t)
\\ w_2Center_x
\\ w_2Center_y
\\ w_2Area
\\ w_3p(C_1)
\\ w_3p(C_2)
\\ \vdots
\\ w_3p(C_n)
\end{bmatrix},
\end{equation}
where $MSE(X_t,\widehat{X}_t)$ is the prediction error from the GAN-based frame predictor (Section \ref{s:predict}); $Center_x, Center_y, Area$ denote the coordinates of the center of the bounding box and the area of the bounding box (Section \ref{s:detect}); and $p(C_1),\ldots,p(C_n)$ are the class probabilities for the detected object (Section \ref{s:detect}). Hence, at any given time $t$, with $n$ denoting the number of possible classes, the dimensionality of $F_t^i$ is given by
$m=n+4$. 

\subsection{Anomaly Detection}
\label{s:anomaly}

Our goal here is to detect anomalies in streaming videos with minimal detection delays while satisfying a desired false alarm rate. We can safely hypothesize that any anomalous event would persist for an unknown period of time. This makes the problem suitable for a sequential anomaly detection framework \cite{basseville1993detection}. However, since we have no prior knowledge about the anomalous event that might occur in a video, parametric algorithms which require probabilistic model and data for both nominal and anomaly cannot be used directly. Next, we explain the training and testing of our proposed nonparametric sequential anomaly detection algorithm.

\textbf{Training:} First, given a set of $N$ training videos ${\mathcal{V} \triangleq \{v_i : i = 1,2,\dots ,N \}}$ consisting of $P$ frames in total, we leverage the deep learning module of our proposed detector to extract $M$ feature vectors $\mathcal{F}^M=\{F^i\}$ for $M$ detected objects in total such that $M \geq P$. We assume that the training data does not include any anomalies. These $M$ vectors correspond to $M$ points in the nominal data space, distributed according to an unknown complex probability distribution. Following a data-driven approach we would like to learn a nonparametric description of the nominal data distribution. Due to its attractive traits, such as analyzability, interpretability, and computational efficiency \cite{chen2018explaining,gu2019statistical}, we use $k$ nearest neighbor ($k$NN) distance, which captures the local interactions between nominal data points, to figure out a nominal data pattern. Given the informativeness of extracted motion, location, and appearance features, anomalous instances are expected to lie further away from the nominal manifold defined by $\mathcal{F}^M$. Consequently, the $k$NN distance of anomalous instances with respect to the nominal data points in $\mathcal{F}^M$ will be statistically higher as compared to the nominal data points. The training procedure of our detector is given as follows:

\begin{enumerate}
\item Randomly partition the nominal dataset $\mathcal{F}^M$ into two sets $\mathcal{F}^{M_1}$ and $\mathcal{F}^{M_2}$ such that $M = M_1 + M_2$. 
\item Then for each point $F_i$ in $\mathcal{F}^{M_1}$, we compute the $k$NN distance $d_i$ with respect to the points in set $\mathcal{F}^{M_2}$. 
\item For a significance level $\alpha$, e.g., $0.05$, the $(1-\alpha)$th percentile $d_\alpha$ of $k$NN distances $\{d_1,\ldots,d_{M_1}\}$ is used as a baseline statistic for computing the anomaly evidence of test instances.
\item The maximum value of $k$NN distances $\{d_1,\ldots,d_{M_1}\}$ is used as an upper bound ($\phi$) for $\delta_t$, given by Eq. \eqref{eq:evidence}, which is then used for selecting a threshold $h$, as explained in Section \ref{s:thr}.
\end{enumerate}

\textbf{Testing:} During the testing phase, for each object $i$ detected at time $t$, the deep learning module constructs the feature vector $F_t^i$ and computes the $k$NN (Euclidean) distance $d_t^i$ with respect to the training instances in $\mathcal{F}^{M_2}$. The proposed sequential anomaly detection system then computes the instantaneous frame-level anomaly evidence $\delta_t$:
\begin{equation}
\label{eq:evidence}
    \delta_t = (\max_i\{d_t^i\})^m - d_\alpha^m,
\end{equation}
where $m$ is the dimensionality of feature vector $F_t^i$. Finally, following a CUSUM-like procedure \cite{basseville1993detection} we update the running decision statistic $s_t$ as
\begin{equation}
    s_t = \max\{s_{t-1} + \delta_t,0\}, s_0 = 0.
\end{equation}
For nominal data, $\delta_t$ typically gets negative values, hence the decision statistic $s_t$ hovers around zero; whereas for anomalous data $\delta_t$ is expected to take positive values, and successive positive values of $\delta_t$ will make $s_t$ grow. 
We decide that a video frame is anomalous if the decision statistic $s_t$ exceeds the threshold $h$. After $s_t$ exceeds $h$, we perform some fine tuning to better label video frames as nominal or anomalous. Specifically, we find the frame $s_t$ started to grow, i.e., the last time $s_t=0$ before detection, say $\tau_{start}$. Then, we also determine the frame $s_t$ stops increasing and keeps decreasing for $n$, e.g., $5$, consecutive frames, say $\tau_{end}$. Finally, we label the frames between $\tau_{start}$ and $\tau_{end}$ as anomalous, and continue testing for new anomalies with frame $\tau_{end}+1$ by resetting $s_{\tau_{end}}=0$.

\subsection{Threshold Selection}
\label{s:thr}

If the test statistic crosses the threshold when there is no anomaly, this event is called a false alarm. Existing works consider the decision threshold as a design parameter, and do not provide any analytical procedure for choosing its value. For an anomaly detection algorithm to be implemented in a practical setting, a clear procedure is necessary for selecting the decision threshold such that it satisfies a desired false alarm rate. The reliability of an algorithm in terms of false alarm rate is crucial for minimizing human involvement. To provide such a performance guarantee for the false alarm rate, we derive an asymptotic upper bound on the average false alarm rate of the proposed algorithm.

\begin{thm}
\label{thm:1}
The false alarm rate of the proposed algorithm is asymptotically (as $M_2 \rightarrow \infty$) upper bounded by 
\begin{equation}
    FAR \leq e^{-\omega_0h}, 
\end{equation}
where $h$ is the decision threshold, and $\omega_0>0$ is given by
\begin{align}
    \label{e:thm}
    \omega_0 &= v_m - \theta -\frac{1}{\phi} \mathcal{W}\left( -\phi \theta e^{-\phi\theta } \right), \\
    \theta &= \frac{v_m}{e^{v_m d_\alpha^m}}.\nonumber
\end{align}
In \eqref{e:thm}, $\mathcal{W}(\cdot)$ is the Lambert-W function, $v_m=\frac{\pi^{m/2}}{\Gamma(m/2+1)}$ is the constant for the $m$-dimensional Lebesgue measure (i.e., $v_m d_\alpha^m$ is the $m$-dimensional volume of the hyperball with radius $d_\alpha$), and $\phi$ is the upper bound for $\delta_t$.
\end{thm}
\textit{Proof.} See Appendix.

Although the expression for $\omega_0$ looks complicated, all the terms in \eqref{e:thm} can be easily computed. Particularly, $v_m$ is directly given by the dimensionality $m$, $d_\alpha$ comes from the training phase, $\phi$ is also found in training, and finally there is a built-in Lambert-W function in popular programming languages such as Python and Matlab. 
Hence, given the training data, $\omega_0$ can be easily computed, and based on Theorem \ref{thm:1}, the threshold $h$ can be chosen to asymptotically achieve the desired false alarm period as follows
\begin{equation}
h = \frac{-\log (FAR)}{\omega_0}.
\end{equation}

\section{Experiments}
\label{experiments}

\subsection{Datasets}

We evaluate our proposed method on three publicly available video anomaly data sets, namely the CUHK avenue dataset \cite{lu2013abnormal}, the UCSD pedestrian dataset \cite{mahadevan2010anomaly}, and the ShanghaiTech \cite{luo2017revisit} campus dataset. Each data set presents its own set of challenges and unique characteristics such as types of anomaly, video quality, background location, etc. Hence, we treat each dataset independently and present individual results for each of them. Here, we briefly introduce each dataset that are used in our experiments.

\textbf{UCSD:} The UCSD pedestrian data set is composed of two parts, namely Ped1 and Ped2. Following the work of \cite{ionescu2019object,hinami2017joint}, we exclude Ped1 from our experiments due to its significantly lower resolution of 158 x 238 and a lack of consistency in the reported results as some recent works reported their performance only on a subset of the entire data set. Hence, we present our results on the UCSD Ped2 dataset which consists of 16 training and 12 test videos, each with a resolution of 240 x 360. All the anomalous events are caused due to vehicles such as bicycles, skateboarders and wheelchairs crossing pedestrian areas. 

\textbf{Avenue:} The CUHK avenue dataset consists of 16 training and 21 test videos with a frame resolution of 360 x 640. The anomalous behaviour is represented by people throwing objects, loitering and running. 

\textbf{ShanghaiTech:} The ShanghaiTech Campus dataset is one of the largest and most challenging datasets available for anomaly detection in videos. It consists of 330 training and 107 test videos from 13 different scenes, which sets it apart from the other available datasets. The resolution for each video frame is 480 x 856.

\subsection{Comparison with Existing Methods}

We compare our proposed algorithm in Table \ref{tab:my-table} with state-of-the-art deep learning-based methods, as well as methods based on hand-crafted features: MPPCA \cite{kim2009observe}, MPPC + SFA \cite{mahadevan2010anomaly}, Del et al. \cite{del2016discriminative}, Conv-AE \cite{hasan2016learning}, ConvLSTM-AE \cite{luo2017remembering}, Growing Gas \cite{sun2017online}, Stacked RNN \cite{luo2017revisit}, Deep Generic \cite{hinami2017joint}, GANs \cite{ravanbakhsh2018plug}, Liu et al. \cite{liu2018future}. %, Sultani et al. \cite{sultani2018real}. 
A popular metric used for comparison in anomaly detection literature is the Area under the Receiver Operating Characteristic (AuROC) curve. Higher AuROC values indicate better performance for an anomaly detection system. For performance evaluation, following the existing works \cite{cong2011sparse,ionescu2019object,liu2018future}, we consider frame level AuROC.

\subsection{Implementation Details}

In the prediction pipeline, the U-NET based generator and the patch discriminator are implemented in Tensorflow. Each frame is resized to 256 x 256 and normalized to [-1,1]. The window size $t$ is set to 4. Similar to \cite{liu2018future}, we use the Adam optimizer for training and set the learning rate to 0.0001 and 0.00001 for the generator and discriminator, respectively. The object detector used is YOLOv3 which is based on the Darknet architecture and is pretrained on the MS-COCO dataset.
During training, we extract the bounds which have a confidence level greater than 0.6, and for testing we consider confidence levels greater than or equal to 0.4. The weights $w_1, w_2$ and $w_3$ are set to 1, 0.4 and 0.9 respectively. The sequential anomaly detection algorithm is implemented in Python.  

\subsection{Impact of Sequential Anomaly Detection}

\begin{figure}[th]
\centering
\includegraphics[width=.7\textwidth]{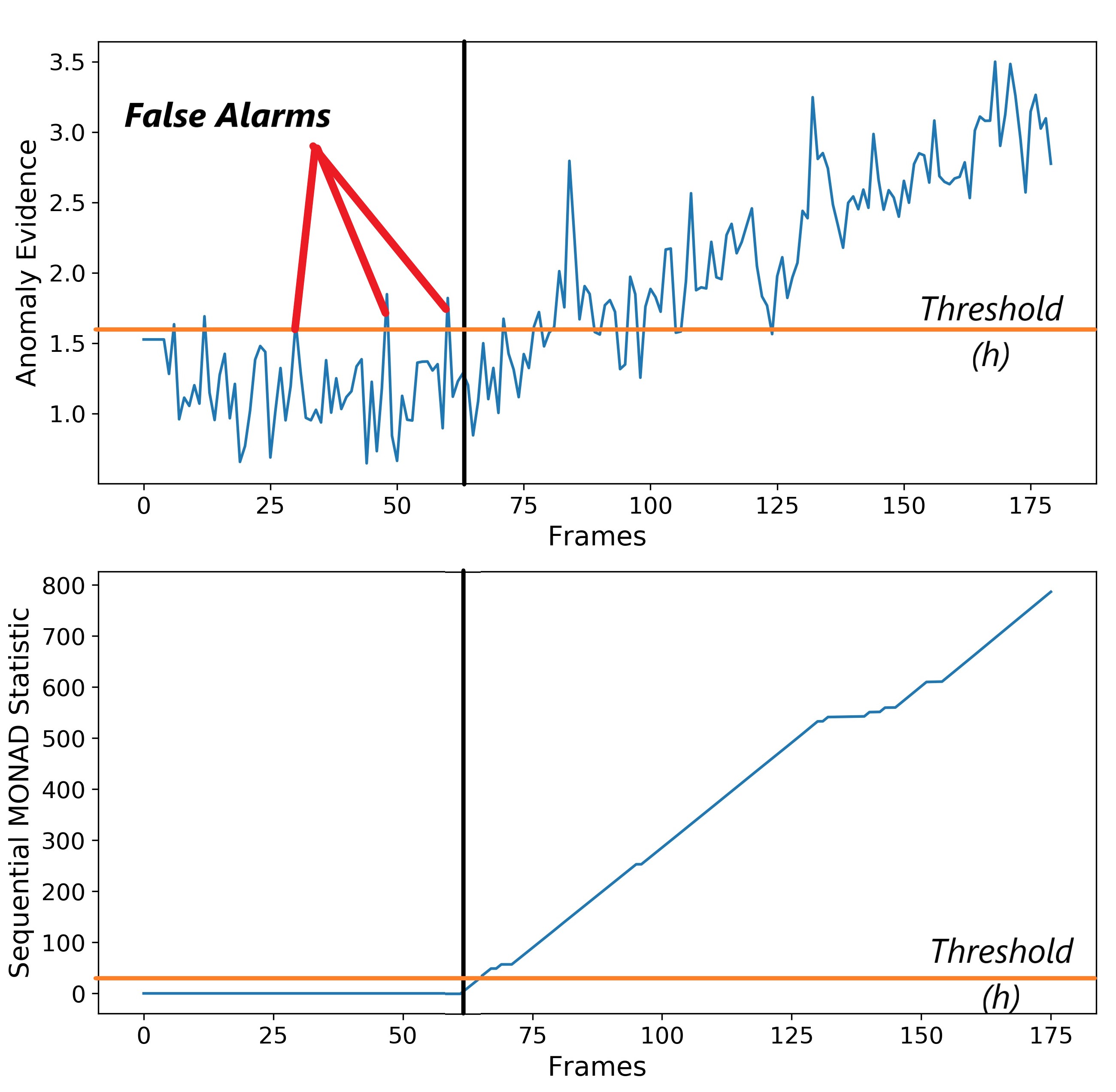}
\vspace{-2mm}
\caption{The advantage of sequential anomaly detection over single-shot detection in terms of controlling false alarms.}
\label{f:sequential}
\vspace{-2mm}
\end{figure}

To demonstrate the importance of sequential anomaly detection in videos, we implement a nonsequential version of our algorithm by applying a threshold to the instantaneous anomaly evidence $\delta_t$, given in \eqref{eq:evidence}, which is similar to the approach employed by many recent works \cite{liu2018future,sultani2018real,ionescu2019object}. As Figure \ref{f:sequential} shows, instantaneous anomaly evidence is more prone to false alarms than the sequential MONAD statistic since it only considers the noisy evidence available at the current time to decide. Whereas, the proposed sequential statistic handles noisy evidence by integrating recent evidence over time. 

\subsection{Results}

We compare our results to a wide range of methods in Table \ref{tab:my-table}. Recently, \cite{ionescu2019object} showed significant gains over the rest of the methods. However, their methodology of computing the AuROC gives them an unfair advantage as they calculate the AuROC for each video in a dataset, and then average them as the AuROC of the dataset, as opposed to the other works which concatenate all the videos first and then determine the AuROC as the dataset's score.
\begin{table}[]
\centering
\resizebox{0.8\textwidth}{!}{%
\begin{tabular}{|c|c|c|c|c|}
\hline
Methodology    & CUHK Avenue & UCSD Ped 2 & ShanghaiTech  \\ \hline
MPPCA \cite{kim2009observe} & -           & 69.3       & -           \\ \hline
MPPC + SFA \cite{mahadevan2010anomaly}     & -           & 61.3       & -   \\ \hline
Del et al. \cite{del2016discriminative}     & 78.3        & -          & -   \\ \hline
Conv-AE \cite{hasan2016learning}        & 80.0        & 85.0       & 60.9\\ \hline
ConvLSTM-AE \cite{luo2017remembering}    & 77.0        & 88.1       & -   \\ \hline
Growing Gas \cite{sun2017online}    & -           & 93.5       & -   \\ \hline
Stacked RNN \cite{luo2017revisit}    & 81.7        & 92.2       & 68.0\\ \hline
Deep Generic \cite{hinami2017joint}   & -           & 92.2       & -   \\ \hline
GANs \cite{ravanbakhsh2017abnormal}           & -           & 88.4       & -   \\ \hline
Liu et al. \cite{liu2018future}     & 85.1        & 95.4       & \textbf{72.8}\\ \hline
\textbf{Ours}           & \textbf{86.4}        & \textbf{97.2}       & 70.9     \\ \hline
\end{tabular}%
}
\caption{AuROC result comparison on three datasets.}
\label{tab:my-table}
\end{table}

As shown in Table \ref{tab:my-table} we are able to outperform the existing results in the avenue and UCSD dataset, and achieve competitive performance in the ShanghaiTech dataset. We should note here that our reported result in the ShanghaiTech dataset is based on online decision making without seeing future video frames.
A common technique used by several recent works \cite{liu2018future,ionescu2019object} is to normalize the computed statistic for each test video independently, including the ShanghaiTech dataset. However, this methodology cannot be implemented in an online (real-time) system as it requires prior knowledge about the minimum and maximum values the statistic might take. 

\begin{figure}[h]
\centering
\includegraphics[width=0.49\textwidth]{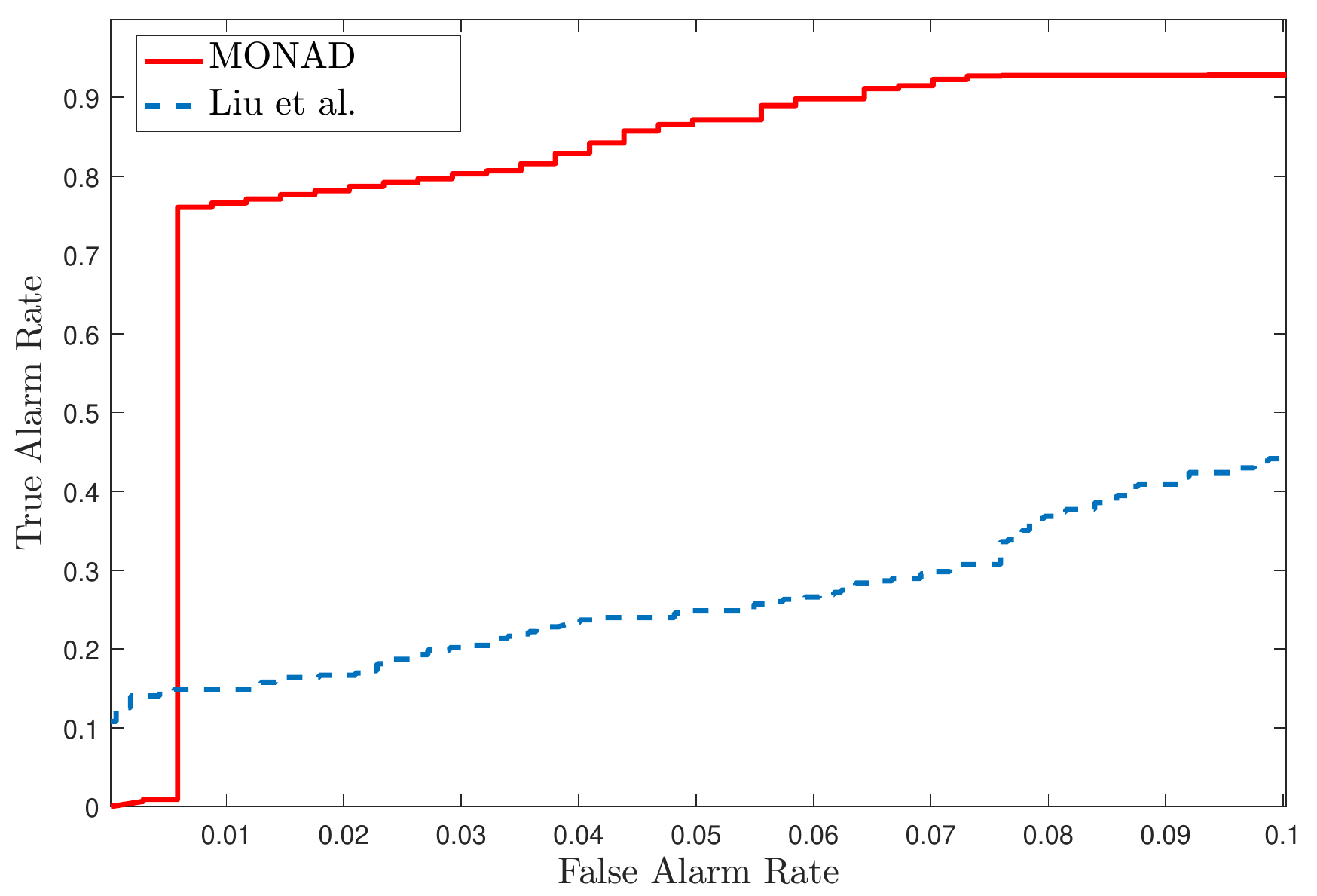}
\includegraphics[width=0.5\textwidth]{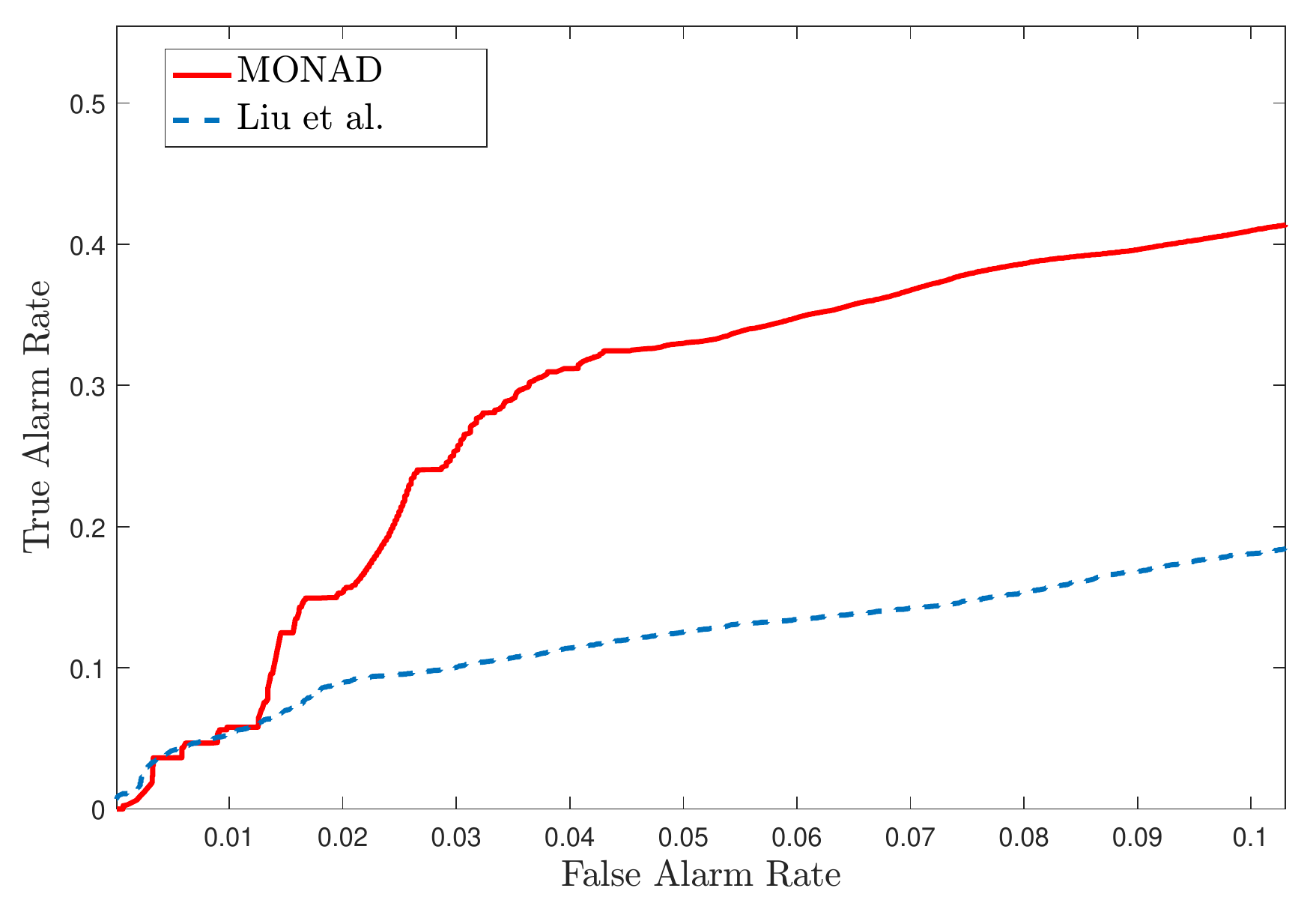}
\vspace{-2mm}
\caption{The ROC curves of the proposed MONAD algorithm and the online version of Liu et al. \cite{liu2018future} for a practical range of false alarm rate in the UCSD Ped 2 (left) and ShanghaiTech (right) data sets.}
\label{f:prac-ucsd}
\vspace{-2mm}
\end{figure}

Hence, we also compare our online method with the online version of state-of-the-art method \cite{liu2018future}. In that version, the minimum and maximum values of decision statistic is obtained from the training data and used for all videos in the test data to normalize the decision statistic, instead of the minimum and maximum values in each test video separately. AuROC value, which is the most common performance metric in the literature, considers the entire range $(0,1)$ of false alarm rates. However, in practice, false alarm rate must satisfy an acceptable level (e.g., up to 10\%). In Figure \ref{f:prac-ucsd}, on the UCSD and ShanghaiTech data sets, we compare our algorithm with the online version of \cite{liu2018future} within a practical range of false alarm in terms of the ROC curve (true positive rate vs. false positive rate). As clearly seen in the figures, the proposed MONAD algorithm achieves much higher true alarm rates than \cite{liu2018future} in both datasets while satisfying practical false alarm rates.

\begin{figure*}[!htb]
\minipage{0.5\textwidth}
  \includegraphics[width=\linewidth]{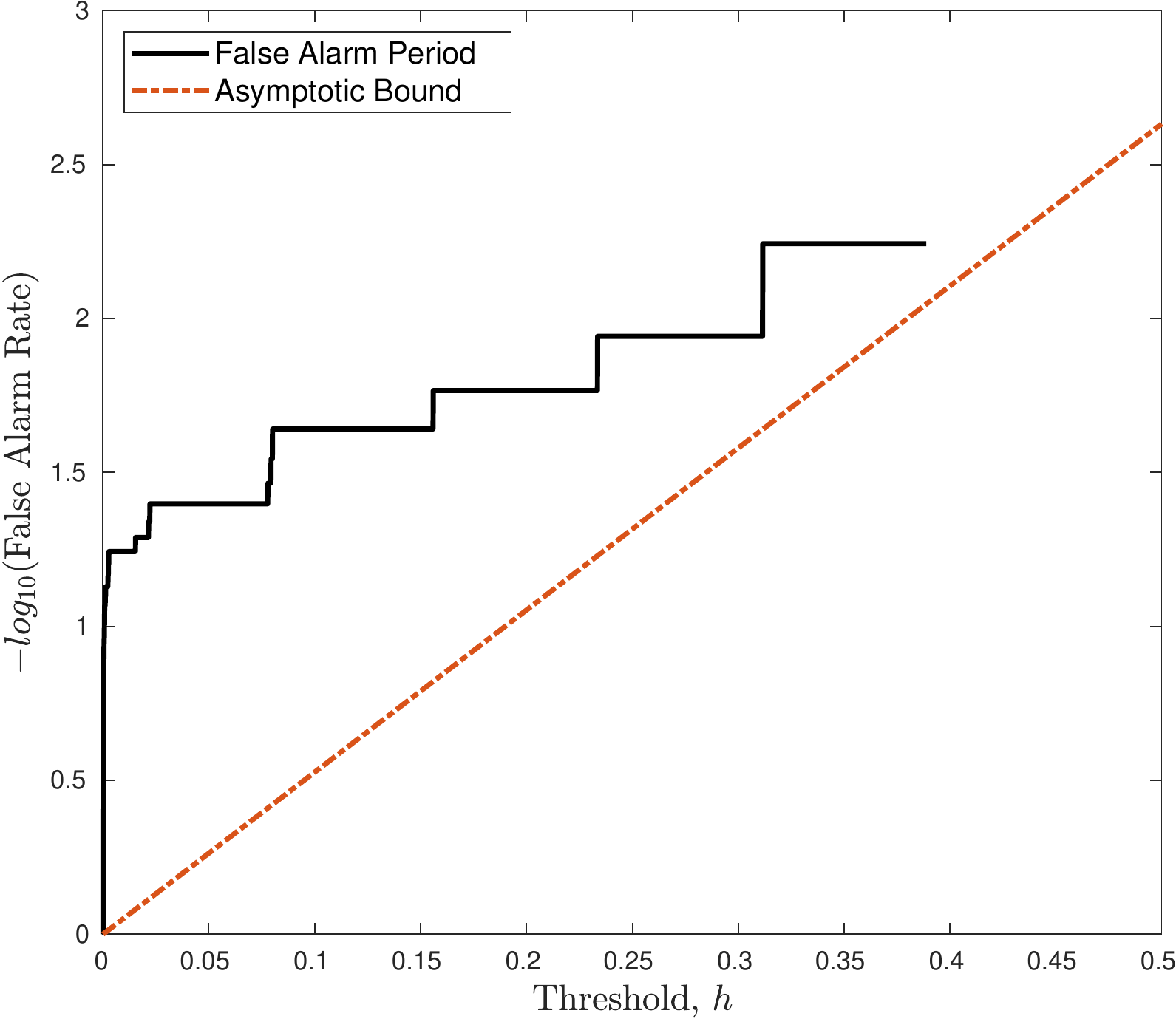}
\label{f:lb_UCSD}
\endminipage\hfill
\minipage{0.5\textwidth}
  \includegraphics[width=\linewidth]{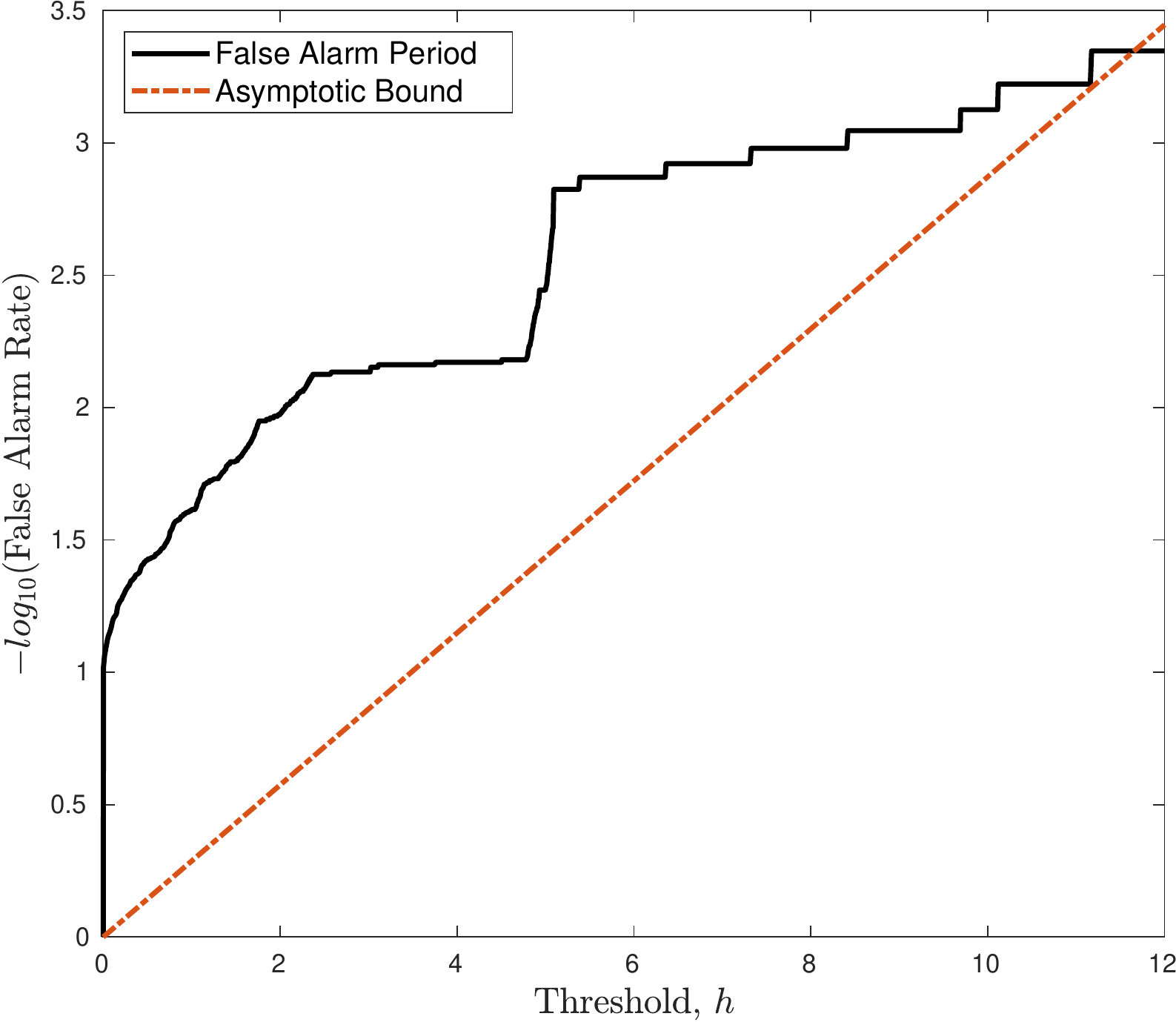}
  \label{f:lb_Shanghai}
\endminipage\hfill
%\minipage{0.5\textwidth}%
  \centering
  \includegraphics[width=.5\textwidth]{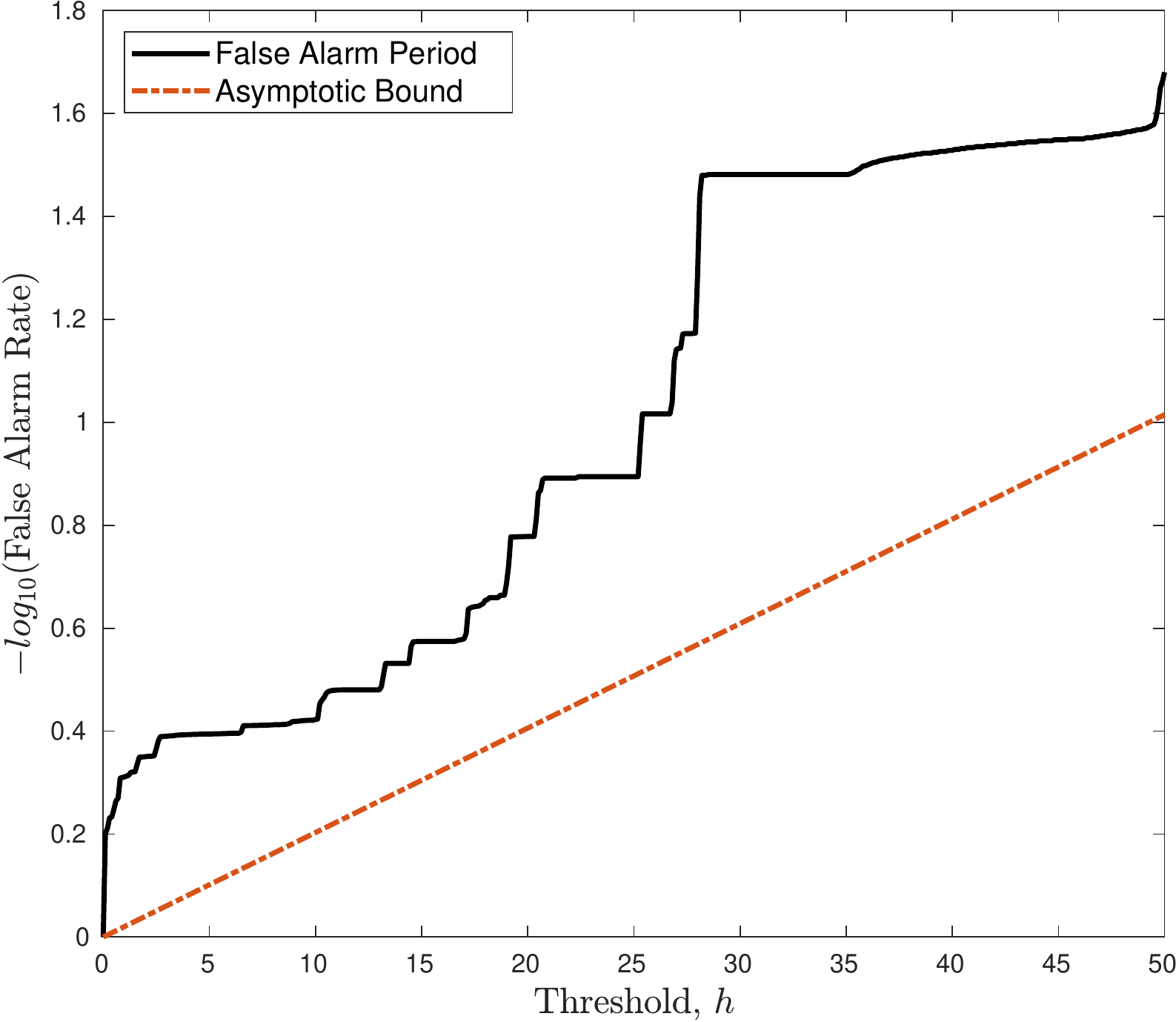}
%  \label{f:lb_Avenue}
%\endminipage
\caption{Actual false alarm periods vs. derived lower bounds for the UCSD Ped.2 (top left), ShanghaiTech (top right), and Avenue (bottom) data sets.}
\label{f:bounds}
\end{figure*}

Finally, in Figure \ref{f:bounds}, we analyze the bound for false alarm rate derived in Theorem \ref{thm:1}. For the clarity of visualization, the figure shows the logarithm of false alarm period, which is the inverse of the false alarm rate. In this case, the upper bound on false alarm rate becomes a lower bound on the false alarm period. The experimental results corroborate the theoretical bound and the procedure presented in Section \ref{s:thr} for obtaining the decision threshold $h$.

\subsection{Computational Complexity}

In this section we analyze the computational complexity of the sequential anomaly detection module, as well as the average running time of the deep learning module.

\textbf{\emph{Sequential Anomaly Detection:}} The training phase of the proposed anomaly detection algorithm requires computation of $k$NN distances for each point in $\mathcal{F}^{M_1}$ to each point in $\mathcal{F}^{M_2}$. Therefore, the time complexity of training phase is given by $\mathcal{O}(M_1M_2 m)$. The space complexity of the training phase is $\mathcal{O}(M_2 m)$ since $M_2$ data instances need to be saved for the testing phase. In the testing phase, since we compute the $k$NN distances of a single point to all data points in $\mathcal{F}^{M_2}$, the time complexity is $\mathcal{O}(M_2 m)$. 

\textbf{\emph{Deep Learning Module:}} The average running time for the GAN-based video frame prediction is 22 frames per second. The YOLO object detector requires about 12 milliseconds to process a single image. This translates to about 83.33 frames per second. The running time can be further improved by using a faster object detector such as YOLOv3-Tiny or a better GPU system. All tests are performed on NVIDIA GeForce RTX 2070 with 8 GB RAM and Intel i7-8700k CPU.

\section{Conclusion}
\label{conclusion}

For video anomaly detection, we presented an online algorithm, called MONAD, which consists of a deep learning-based feature extraction module and a statistical decision making module. The first module is a novel feature extraction technique that combines GAN-based frame prediction and a lightweight object detector. The second module is a sequential anomaly detector, which enables performance analysis. The asymptotic false alarm rate of MONAD is analyzed, and a practical procedure is provided for selecting its detection threshold to satisfy a desired false alarm rate. Through real data experiments, MONAD is shown to outperform the state-of-the-art methods, and yield false alarm rates consistent with the derived asymptotic bounds. For future work, we plan to focus on the importance of timely detection in video \cite{mao2019delay} by proposing a new metric based on the average delay and precision. %and evaluating the performance in terms of , in addition to the commonly used metrics such as true positive rate, false positive rate, and AUC.

\section{Acknowledgements}
This research is funded in part by the Florida Center for Cybersecurity and in part by the U.S. National Science Foundation under the grant \#2029875.

\appendix
\section{Proof of Theorem 1}
In \cite{basseville1993detection}[page 177], for CUSUM-like algorithms with independent increments, such as MONAD with independent $\delta_t$, a lower bound on the average false alarm period is given as follows 
\[
E_\infty[T] \geq e^{\omega_0h},
\]
where $h$ is the detection threshold, and $\omega_0 \geq 0$ is the solution to $E[e^{\omega_0\delta_t}] = 1$. 

To analyze the false alarm period, we need to consider the nominal case. In that case, since there is no anomalous object at each time $t$, the selection of object with maximum $k$NN distance in $\delta_t=(\max_i\{d_t^i\})^m-d_\alpha^m$ does not necessarily depend on the previous selections due to lack of an anomaly which could correlate the selections. Hence, in the nominal case, it is safe to assume that $\delta_t$ is independent over time.

We firstly derive the asymptotic distribution of the frame-level anomaly evidence $\delta_t$ in the absence of anomalies. Its cumulative distribution function is given by 
\[
P(\delta_t \leq y) = P((\max_i\{d_t^i\})^m \leq d_\alpha^m + y).
\]
It is sufficient to find the probability distribution of $(\max\limits_i\{d_t^i\})^m$, the $m$th power of the maximum $k$NN distance among objects detected at time $t$. As discussed above, choosing the object with maximum distance in the absence of anomaly yields independent $m$-dimensional instances $\{F_t\}$ over time, which form a Poisson point process. The nearest neighbor ($k=1$) distribution for a Poisson point process is given by
\[
P(\max_i\{d_t^i\} \leq r) = 1 - \exp(-\Lambda(b(F_t,r)))
\]
where $\Lambda(b(F_t,r))$ is the arrival intensity (i.e., Poisson rate measure) in the $m$-dimensional hypersphere $b(F_t,r)$ centered at $F_t$ with radius $r$ \cite{chiu2013stochastic}. Asymptotically, for a large number of training instances as $M_2\to\infty$, under the null (nominal) hypothesis, the nearest neighbor distance $\max_i\{d_t^i\}$ of $F_t$ takes small values, defining an infinitesimal hyperball with homogeneous intensity $\lambda=1$ around $F_t$. Since for a homogeneous Poisson process the intensity is written as $\Lambda(b(F_t,r)) = \lambda |b(F_t,r)|$ \cite{chiu2013stochastic}, where $|b(F_t,r)| = \frac{\pi^{m/2}}{\Gamma(m/2+1)}r^m = v_m r^m$ is the Lebesgue measure (i.e., $m$-dimensional volume) of the hyperball $b(F_t,r)$, we rewrite the nearest neighbor distribution as
\begin{align*}
P(\max_i\{d_t^i\} \le r) &= 1-\exp\left( -v_m r^m \right),
\end{align*}
where $v_m = \frac{\pi^{m/2}}{\Gamma(m/2+1)}$ is the constant for the $m$-dimensional Lebesgue measure. 

Now, applying a change of variables we can write the probability density of $(\max_i\{d_t^i\})^m$ and $\delta_t$ as
\begin{align}
f_{(\max_i\{d_t^i\})^m}(y) &= \frac{\partial}{\partial y} \left[1-\exp\left( -v_m y \right) \right], \\
&= v_m \exp(-v_m y), \\
\label{eq:pdf}
f_{\delta_t}(y) &= v_m \exp(-v_m d_{\alpha}^m) \exp(-v_m y)
\end{align}

Using the probability density derived in \eqref{eq:pdf}, $E[e^{\omega_0\delta_t}] = 1$ can be written as
\begin{align}
1 &= \int_{-d_\alpha^m}^{\phi} e^{\omega_0 y}v_me^{-v_m d_\alpha^m}e^{-v_my}dy, \\
\frac{e^{v_md_\alpha^m}}{v_m} &= \int_{-d_\alpha^m}^{\phi} e^{(\omega_0-v_m)y}dy, \\
&= \frac{e^{(\omega_0-v_m)y}}{\omega_0-v_m}\Biggr|_{-d_\alpha^m}^{\phi}, \\ 
&=\frac{e^{(\omega_0-v_m)\phi} - e^{(\omega_0-v_m)(-d_\alpha^m)}}{\omega_0-v_m},
\end{align}
where $-d_\alpha^m$ and $\phi$ are the lower and upper bounds for $\delta_t=(\max_i\{d_t^i\})^m-d_\alpha^m$. The upper bound $\phi$ is obtained from the training set. 

As $M_2\to\infty$, since the $m$th power of $(1-\alpha)$th percentile of nearest neighbor distances in training set goes to zero, i.e., $d_\alpha^m \to 0$, we have 
\begin{align}
e^{(\omega_0-v_m)\phi} &= \frac{e^{v_m d_\alpha^m}}{v_m}(\omega_0-v_m) + 1. 
\end{align}

We next rearrange the terms to obtain the form of $e^{\phi x} = a_0(x+\theta)$ where $x=\omega_0-v_m$, $a_0=\frac{e^{v_m d_\alpha^m}}{v_m}$, and $\theta=\frac{v_m}{e^{v_m d_\alpha^m}}$. The solution for $x$ is given by the Lambert-W function \cite{scott2014asymptotic} as $x = -\theta - \frac{1}{\phi} \mathcal{W}(-\phi e^{-\phi\theta}/a_0)$, hence
\begin{align}
\omega_0 = v_m - \theta -\frac{1}{\phi} \mathcal{W}\left( -\phi \theta e^{-\phi\theta } \right). 
\end{align}

Finally, since the false alarm rate (i.e., frequency) is the inverse of false alarm period $E_\infty[T]$, we have 
\[
FAR \leq e^{-\omega_0 h},
\]
where $h$ is the detection threshold, and $\omega_0$ is given above.
%\section{Further Comparisons in Practical Setups}
%\input{files/further_comp.tex}

\bibliography{example_paper}

\newpage
\textbf{Keval Doshi} received the B.Sc. degree in Electronics and Communications Engineering from Gujarat Technological University, India, in 2017. He is currently a Ph.D. student at the Electrical Engineering Department at the University of South Florida, Tampa. His research interests include computer vision, machine learning and cybersecurity.

\textbf{Yasin Yilmaz} received the Ph.D. degree in Electrical Engineering from Columbia University, New York, NY, in 2014. He is currently an Assistant Professor of Electrical Engineering at the University of South Florida, Tampa. He received the Collaborative Research Award from Columbia University in 2015. His research interests include machine learning, statistical signal processing, and their applications to computer vision, cybersecurity, IoT networks, energy systems, transportation systems, environmental systems, and communication systems.

\end{document}

\twocolumn[
\icmltitle{[Supplementary File] Online Anomaly Detection in Surveillance Videos with Asymptotic Bounds on False Alarm Rate}
]

\section*{Corrections}

\begin{itemize}
    \item ``Theorem 1" before eq. (11)
    \item ``Figure 4" in the last paragraph of Section 4.5
\end{itemize}
The algorithms in Table I are MPPCA \cite{kim2009observe}, MPPC + SFA \cite{mahadevan2010anomaly}, Del et al. \cite{del2016discriminative}, Conv-AE \cite{hasan2016learning}, ConvLSTM-AE \cite{luo2017remembering}, Growing Gas \cite{sun2017online}, Stacked RNN \cite{luo2017revisit}, Deep Generic \cite{hinami2017joint}, GANs \cite{ravanbakhsh2018plug}, Liu et al. \cite{liu2018future}, Sultani et al. \cite{sultani2018real}. 

We noticed that the submitted version lacks the Conclusion section. Please see it below. 

\textbf{\textit{Conclusion:}} For video anomaly detection, we presented an online algorithm, called MONAD, which consists of a deep learning-based feature extraction module and a statistical decision making module. The first module is a novel feature extraction technique that combines GAN-based frame prediction and a lightweight object detector. The second module is a sequential anomaly detector, which enables performance analysis. The asymptotic false alarm rate of MONAD is analyzed, and a practical procedure is provided for selecting its detection threshold to satisfy a desired false alarm rate. Through real data experiments, MONAD is shown to outperform the state-of-the-art methods, and yield false alarm rates similar to the derived asymptotic lower bounds.

\section*{Further Comparisons in Practical Setups}

In a practical surveillance setting, an anomaly detector should operate in an online fashion without requiring future video frames. Since most of the reported results in the literature, including the state-of-the-art method \cite{liu2018future}, assume the availability of future video frames, we would like to compare our online method with the online version of state-of-the-art method \cite{liu2018future}. In that version, the minimum and maximum values of decision statistic is obtained from the training data and used for all videos in the test data to normalize the decision statistic, instead of the minimum and maximum values in each test video separately. Moreover, AuROC value, which is the most common performance metric in the literature, considers the entire range $(0,1)$ of false alarm rates. However, in practice, false alarm rate must satisfy an acceptable level (e.g., up to 10\%). In Figures \ref{f:prac-ucsd} and \ref{f:prac-shang}, we compare our algorithm with the online version of \cite{liu2018future} within a practical range of false alarm in terms of the true alarm rate (i.e., detection probability). As clearly seen in the figures, the proposed MONAD algorithm achieves much higher true alarm rates than \cite{liu2018future} in both datasets while satisfying practical false alarm rates.

\begin{figure}[h]
\centering
\includegraphics[width=0.5\textwidth]{Images/global_norm_ucsd_zoom.pdf}
\vspace{-2mm}
\caption{The ROC curves of online algorithms for a practical range of false alarm rate in the UCSD Ped 2 dataset.}
\label{f:prac-ucsd}
\vspace{-2mm}
\end{figure}

\begin{figure}[h]
\centering
\includegraphics[width=0.5\textwidth]{Images/global_norm_shang_zoom.pdf}
\vspace{-2mm}
\caption{The ROC curves of online algorithms for a practical range of false alarm rate in the ShanghaiTech dataset.}
\label{f:prac-shang}
\vspace{-2mm}
\end{figure}

\section*{Proof of Theorem 1}

In \cite{basseville1993detection}[page 177], for CUSUM-like algorithms with independent increments, such as MONAD with independent $\delta_t$, a lower bound on the average false alarm period is given as follows 
\[
E_\infty[T] \geq e^{\omega_0h},
\]
where $h$ is the detection threshold, and $\omega_0 \geq 0$ is the solution to $E[e^{\omega_0\delta_t}] = 1$. 

To analyze the false alarm period, we need to consider the nominal case. In that case, since there is no anomalous object at each time $t$, the selection of object with maximum $k$NN distance in $\delta_t=(\max_i\{d_t^i\})^m-d_\alpha^m$ does not necessarily depend on the previous selections due to lack of an anomaly which could correlate the selections. Hence, in the nominal case, it is safe to assume that $\delta_t$ is independent over time.

We firstly derive the asymptotic distribution of the frame-level anomaly evidence $\delta_t$ in the absence of anomalies. Its cumulative distribution function is given by 
\[
P(\delta_t \leq y) = P((\max_i\{d_t^i\})^m \leq d_\alpha^m + y).
\]
It is sufficient to find the probability distribution of $(\max\limits_i\{d_t^i\})^m$, the $m$th power of the maximum $k$NN distance among objects detected at time $t$. As discussed above, choosing the object with maximum distance in the absence of anomaly yields independent $m$-dimensional instances $\{F_t\}$ over time, which form a Poisson point process. The nearest neighbor ($k=1$) distribution for a Poisson point process is given by
\[
P(\max_i\{d_t^i\} \leq r) = 1 - \exp(-\Lambda(b(F_t,r)))
\]
where $\Lambda(b(F_t,r))$ is the arrival intensity (i.e., Poisson rate measure) in the $m$-dimensional hypersphere $b(F_t,r)$ centered at $F_t$ with radius $r$ \cite{chiu2013stochastic}. Asymptotically, for a large number of training instances as $M_2\to\infty$, under the null (nominal) hypothesis, the nearest neighbor distance $\max_i\{d_t^i\}$ of $F_t$ takes small values, defining an infinitesimal hyperball with homogeneous intensity $\lambda=1$ around $F_t$. Since for a homogeneous Poisson process the intensity is written as $\Lambda(b(F_t,r)) = \lambda |b(F_t,r)|$ \cite{chiu2013stochastic}, where $|b(F_t,r)| = \frac{\pi^{m/2}}{\Gamma(m/2+1)}r^m = v_m r^m$ is the Lebesgue measure (i.e., $m$-dimensional volume) of the hyperball $b(F_t,r)$, we rewrite the nearest neighbor distribution as
\begin{align*}
P(\max_i\{d_t^i\} \le r) &= 1-\exp\left( -v_m r^m \right),
\end{align*}
where $v_m = \frac{\pi^{m/2}}{\Gamma(m/2+1)}$ is the constant for the $m$-dimensional Lebesgue measure. 

Now, applying a change of variables we can write the probability density of $(\max_i\{d_t^i\})^m$ and $\delta_t$ as
\begin{align}
f_{(\max_i\{d_t^i\})^m}(y) &= \frac{\partial}{\partial y} \left[1-\exp\left( -v_m y \right) \right], \nn\\
&= v_m \exp(-v_m y), \nn\\
\label{eq:pdf}
f_{\delta_t}(y) &= v_m \exp(-v_m d_{\alpha}^m) \exp(-v_m y)
\end{align}

Using the probability density derived in \eqref{eq:pdf}, $E[e^{\omega_0\delta_t}] = 1$ can be written as
\begin{align}
1 &= \int_{-d_\alpha^m}^{\phi} e^{\omega_0 y}v_me^{-v_m d_\alpha^m}e^{-v_my}dy, \nn\\
\frac{e^{v_md_\alpha^m}}{v_m} &= \int_{-d_\alpha^m}^{\phi} e^{(\omega_0-v_m)y}dy, \nn\\
&= \frac{e^{(\omega_0-v_m)y}}{\omega_0-v_m}\Biggr|_{-d_\alpha^m}^{\phi}, \nn\\ 
&=\frac{e^{(\omega_0-v_m)\phi} - e^{(\omega_0-v_m)(-d_\alpha^m)}}{\omega_0-v_m},
\end{align}
where $-d_\alpha^m$ and $\phi$ are the lower and upper bounds for$\delta_t=(\max_i\{d_t^i\})^m-d_\alpha^m$. The upper bound $\phi$ is obtained from the training set. 

As $M_2\to\infty$, since the $m$th power of $(1-\alpha)$th percentile of nearest neighbor distances in training set goes to zero, i.e., $d_\alpha^m \to 0$, we have 
\begin{align}
e^{(\omega_0-v_m)\phi} &= \frac{e^{v_m d_\alpha^m}}{v_m}(\omega_0-v_m) + 1. \nn
\end{align}

We next rearrange the terms to obtain the form of $e^{\phi x} = a_0(x+\theta)$ where $x=\omega_0-v_m$, $a_0=\frac{e^{v_m d_\alpha^m}}{v_m}$, and $\theta=\frac{v_m}{e^{v_m d_\alpha^m}}$. The solution for $x$ is given by the Lambert-W function \cite{scott2014asymptotic} as $x = -\theta - \frac{1}{\phi} \mathcal{W}(-\phi e^{-\phi\theta}/a_0)$, hence
\begin{align}
\omega_0 = v_m - \theta -\frac{1}{\phi} \mathcal{W}\left( -\phi \theta e^{-\phi\theta } \right). \nn
\end{align}

Finally, since the false alarm rate (i.e., frequency) is the inverse of false alarm period $E_\infty[T]$, we have 
\[
FAR \leq e^{-\omega_0 h},
\]
where $h$ is the detection threshold, and $\omega_0$ is given above.

\bibliographystyle{icml2020}
\bibliography{ref2}